\pdfoutput=1
\documentclass{article} % For LaTeX2e
\usepackage{iclr2016_conference,times}
\usepackage{hyperref}
\hypersetup{
     colorlinks   = true,
     citecolor    = black
}
\usepackage[linesnumbered]{algorithm2e}
\usepackage{url}
\usepackage{verbatim}
\usepackage[pdftex]{graphicx}
\usepackage{amssymb}
\usepackage{amsmath}
\usepackage{tikz}
\usepackage{graphicx}
\usepackage{caption}
\usepackage{subcaption}

\SetCommentSty{mycommfont}

\title{Conditional computation in neural networks for faster models}

\author{Emmanuel Bengio, Pierre-Luc Bacon, Joelle Pineau \& Doina Precup \\
School of Computer Science\\
McGill University\\
Montreal, Canada \\
\texttt{\{ebengi,pbacon,jpineau,dprecup\}@cs.mcgill.ca}}

\begin{document}

\maketitle

\begin{abstract}
Deep learning has become the state-of-art tool in many applications, but the evaluation and training of deep models can be time-consuming and computationally expensive. 
The conditional computation approach has been proposed to tackle this problem \citep{bengio2013estimating,davis2013low}.  It operates by selectively activating only parts of the network at a time. In this paper, we use reinforcement learning as a tool to optimize conditional computation policies.
More specifically, we cast the problem of learning activation-dependent policies for dropping out blocks of units as a reinforcement learning problem.
We propose a learning scheme motivated by computation speed, capturing the idea of wanting to have parsimonious activations while maintaining prediction accuracy. We apply a policy gradient algorithm for learning policies that optimize this loss function and propose a regularization mechanism that encourages diversification of the dropout policy. We present encouraging empirical results showing that this approach improves the speed of computation without impacting the quality of the approximation.
\end{abstract}


\section{Introduction}
Large-scale neural networks, and in particular deep learning architectures, have seen a surge in popularity in recent years, due to their impressive empirical performance in complex supervised learning tasks, including state-of-the-art performance in image and speech recognition \citep{he2015delving}.   Yet the task of training such networks remains a challenging optimization problem.  Several related problems arise: very long training time (several weeks on modern computers, for some problems), potential for over-fitting (whereby the learned function is too specific to the training data and generalizes poorly to unseen data), and more technically, the vanishing gradient  problem~\citep{Hochreiter91-small,Bengio-et-al-TNN1994}, whereby the gradient information gets increasingly diffuse as it propagates from layer to layer.

 Recent approaches \citep{bengio2013estimating,davis2013low} have proposed the use of {\em conditional computation} in order to address this problem.  Conditional computation refers to activating only some of the units in a network, in an input-dependent fashion. For example, if we think we're looking at a car, we only need to compute the activations of the vehicle detecting units, not of all features that a network could possible compute.  The immediate effect of activating fewer units is that propagating information through the network will be faster, both at training as well as at test time.  However, one needs to be able to decide in an intelligent fashion which units to turn on and off, depending on the input data.  This is typically achieved with some form of gating structure, learned in parallel with the original network.

A secondary effect of conditional computation is that during training, information will be propagated along fewer links.  Intuitively, this allows sharper gradients on the links that do get activated.  Moreover, because only parts of the network are active, and fewer parameters are used in the computation, the net effect can be viewed as a form of regularization of the main network, as the approximator has to use only a small fraction of the possible parameters in order to produce an action.

In this paper, we explore the formulation of conditional computation using reinforcement learning.
We propose to learn input-dependent activation probabilities for every node (or blocks of nodes), while trying to jointly minimize the prediction errors at the output and the number of participating nodes at every layer, thus reducing the computational load.  One can also think of our method as being related to standard dropout, which has been used as a tool to both regularize and speed up the computation.  However, we emphasize that dropout is in fact a form of ``unconditional" computation, in which the computation paths are data-independent.  Therefore, usual dropout is less likely to lead to specialized computation paths within a network.

We present the problem formulation, and our solution to the proposed optimization problem, using policy search methods \citep{Deisenroth2013}.  Preliminary results are included for standard classification benchmarks.

\section{Problem formulation}

Our model consists in a typical fully-connected neural network model, joined with stochastic per-layer policies that activate or deactivate nodes  of the neural network in an input-dependent manner, both at train and test time. The exact algorithm is detailed in appendix \ref{algorithm-pseudocode}.
% by multiplying them by 0 or 1 (i.e. the sampled policy).
%\subsection{Markov Decision Process}

We cast the problem of learning the input-dependent activation probabilities at each layer in the
framework of Markov Decision Processes (MDP) \citep{Puterman1994}. We define a discrete time, continuous state and discrete action MDP $\langle \mathcal{S}, \mathcal{U}, P\left( \cdot \,\middle|\, s, u\right), C\rangle$. % with $C$ the cost function. % and $P\left( \cdot \,\middle|\, s, u\right)$ the distribution over the next state given that action $u$ is taken in state $s$.
An action $\mathbf{u} \in \{0,1\}^k$ in this model consists in the application of a mask over the units of a given layer. We define the state space of the MDP over the vector-valued activations $\mathbf{s}\in \mathbb{R}^k$ of all nodes at the previous layer. The cost $C$ is the loss of the neural network architecture (in our case the negative log-likelihood). This MDP is single-step: an input is seen, an action is taken, a reward is observed and we are at the end state.

Similarly to the way dropout is described \citep{Hinton2012},
each node or block in a given layer has an associated  Bernoulli distribution which determines its probability of being activated.  We train a different policy for each layer $l$, and parameterize it (separately of the neural network) such that it is input-dependent.
For every layer $l$ of $k$ units, we define a policy as a $k$-dimensional Bernoulli distribution:
\begin{align}
\pi^{(l)}(\mathbf{u} \,|\, \mathbf{s}) &= \prod_{i=1}^k \sigma_i^{u_i} (1 - \sigma_i)^{(1 - u_i)}, \hspace{10mm} \mathbf{\sigma}_i = [\text{sigm}(\mathbf{Z}^{(l)}\mathbf{s} + \mathbf{d}^{(l)})]_i, \label{Bernoulli-distrib}
\end{align}
where the $\sigma_i$ denotes the \textit{participation} probability, to be computed from the activations $\mathbf{s}$ of the layer below and the parameters $\theta_l=\{\mathbf{Z}^{(l)}, \mathbf{d}^{(l)}\}$. We denote the sigmoid function by $\text{sigm}$, the weight matrix by $\mathbf{Z}$, and the bias
vector by $\mathbf{d}$. The output of a typical hidden layer $h(x)$ that uses this policy is multiplied element-wise with the mask $\mathbf{u}$ sampled from the probabilities $\mathbf{\sigma}$, and becomes $(h(x)\otimes \mathbf{u})$. For clarity we did not superscript $\mathbf{u}$, $\mathbf{s}$ and $\sigma_i$ with $l$, but each layer has its own. % say better?

%%%%%%%%%%%%%%%%%%%%%%%

\section{Learning sigmoid-Bernoulli policies}
We use REINFORCE \citep{Williams1992} (detailed in appendix \ref{REINFORCE}) to learn the parameters  $\Theta_\pi=\{\theta_1,...,\theta_L\}$ of the sigmoid-Bernoulli policies. Since the nature of the observation space changes at each decision step, we learn $L$ disjoint policies (one for each layer $l$ of the deep network). As a consequence, the summation in the policy gradient disappears and becomes:
\begin{equation}
\nabla_{\theta_l} \mathcal{L} = \mathbb{E}\left\{ C(\mathbf{x}) \nabla_{\theta_l} \log \pi^{(l)}(\mathbf{u}^{(l)} \,|\, \mathbf{s}^{(l)} )\right\} \label{eq:grad-loss-sigm}
\end{equation}
since $\theta_l=\{\mathbf{Z}^{(l)},\mathbf{d}^{(l)}\}$ only appears in the $l$-th decision stage and the gradient is zero otherwise.

Estimating \eqref{eq:grad-loss-sigm} from samples requires propagating through many instances at a
time, which we achieve through mini-batches of size $m_b$ . %This approach has the advantage of making optimal use of the fast matrix-matrix capabilities of modern hardware.
Under the mini-batch
setting, $\mathbf{s}^{(l)}$ becomes a matrix and $\pi(\cdot \,|\, \cdot)$ a vector of dimension
$m_b$ . Taking the gradient of the parameters with respect to the log action probabilities can then
be seen as forming a Jacobian. We can thus re-write the empirical average in matrix form:
\begin{equation}
\nabla_{\theta_l} \mathcal{L}
\approx \frac{1}{m_b} \sum_{i=1}^{m_b} C(\mathbf{x}_i)\nabla_{\theta_l} \log \pi^{(l)}(\mathbf{u}_{i}^{(l)} \,|\, \mathbf{s}_{i}^{(l)}) %\notag
= \frac{1}{m_b} \mathbf{c}^\top \nabla_{\theta_l} \log \pi^{(l)}(\mathbf{U}^{(l)} \,|\, \mathbf{S}^{(l)})
\label{eq:jacobian}
\end{equation}
where $C(\mathbf{x}_i)$ is the total cost for input $\mathbf{x}_i$ and $m_b$ is the number of examples in the mini-batch. The term $\mathbf{c}^\top$ denotes the row vector containing the total costs for every example in the mini-batch. 

%%%%%%%%%%%%%%%%%%%%%%%

\subsection{Fast vector-Jacobian multiplication}
\label{Rop}
While Eqn~\eqref{eq:jacobian} suggests that the Jacobian might have to be formed explicitly,
\citet{Pearlmutter1994} showed that computing a differential derivative suffices to compute left or
right vector-Jacobian (or Hessian) multiplication. The same trick has also recently been revived with
the class of so-called ``Hessian-free'' \citep{Martens2010} methods for artificial neural
networks. Using the notation of \citet{Pearlmutter1994}, we write $\mathcal{R}_{\theta_l}\left\{ \cdot\right\} = \mathbf{c}^\top \nabla_{\theta_l}$ for the differential operator.
\begin{equation}
\nabla_{\theta_l} \mathcal{L} \approx \frac{1}{m_b} \mathcal{R}_{\theta_l}\left\{ \log \pi( \mathbf{U}^{(l)} \,|\, \mathbf{S}^{(l)} )\right\} \label{eq:rop}
\end{equation}

%%%%%%%%%%%%%%%%%%%%%%%
\subsection{Sparsity and variance regularizations}
\label{sec:sparsity}
In order to favour activation policies with \emph{sparse} actions, we add two penalty terms $L_b$ and $L_e$ that depend on some target sparsity rate $\tau$. The first term pushes the policy distribution $\pi$ to activate each unit with probability $\tau$ in expectation over the data.
The second term pushes the policy distribution to have the desired sparsity of activations for each example.
Thus, for a low $\tau$, a valid configuration would be to learn a few high probability activations for some part of the data and low probability activations for the rest of the data, which results in having activation probability $\tau$ in expectation.
\begin{equation}
  L_b = \sum_j^{n}\|\mathbb{E}\{\sigma_{j}\} - \tau\|_2 \quad\quad L_e = \mathbb{E}\{\|(\frac{1}{n}\sum_j^{n} \sigma_{j}) - \tau\|_2\}
\end{equation}
Since we are in a minibatch setting, these expectations can be approximated over the minibatch:
\begin{equation}
  L_b \approx \sum_j^n\|\frac{1}{m_b}\sum_i^{m_b}(\sigma_{ij}) - \tau\|_2 \quad\quad
  L_e \approx \frac{1}{m_b}\sum_i^{m_b}\|(\frac{1}{n}\sum_j^{n} \sigma_{ij}) - \tau\|_2 
\end{equation}
We finally add a third term, $L_v$, in order to favour the aforementioned configurations, where units only have a high probability of activation for certain examples, and low for the rest. We aim to maximize the variances of activations of each unit, across the data. This encourages units' activations to be varied, and while similar in spirit to the $L_b$ term, this term explicitly discourages learning a uniform distribution.
\begin{align}
  L_v &= -\sum_j^n \mbox{var}_i\{\sigma_{ij}\}
  \approx -\sum_j^n \frac{1}{m_b}\sum_i^{m_b} \left(\sigma_{ij} - \left(\frac{1}{m_b}\sum_i^{m_b}\sigma_{ij}\right)\right)^2
\end{align}

%%%%%%%%%%%%%%%%%%%%%%%

\subsection{Algorithm}
\label{sec:algo}

We interleave the learning of the network parameters $\Theta_{NN}$ and the learning of the policy parameters $\Theta_\pi$. We first update the network \emph{and} policy parameters to minimize the following regularized loss function via backpropagation \citep{rumelhart1988learning}:
\begin{align*}
  \mathcal{L} &= -\log P(\mathbf{Y} \,|\, \mathbf{X}, \Theta_{NN}) + \lambda_s (L_b + L_e) + \lambda_v (L_v) + \lambda_{L2}\|\Theta_{NN}\|^2 + \lambda_{L2}\|\Theta_\pi\|^2
\end{align*}
where $\lambda_s$ can be understood as a trade-off parameter between prediction accuracy and parsimony of computation (obtained through sparse node activation), and $\lambda_v$ as a trade-off parameter between a stochastic policy and a more input dependent saturated policy. 
We then minimize the cost function $C$ with a REINFORCE-style approach to update the policy parameters ~\citep{Williams1992}:
\begin{align*}
C = -\log P(\mathbf{Y} \,|\, \mathbf{X}, \Theta_{NN})
\end{align*}
As previously mentioned, we use minibatch stochastic gradient descent as well as minibatch policy gradient updates. A detailed algorithm is available in appendix \ref{algorithm-pseudocode}.

\subsection{Block activation policy}
\label{sub:blockdrop}
To achieve computational gain, instead of activating single units in hidden layers, we activate contiguous (equally-sized) groups of units together (\emph{independently} for each example in the minibatch), thus reducing the action space as well as the number of probabilities to compute and sample.\\
As such, there are two potential speedups. First, the policy is much smaller and faster to compute. Second, it offers a computational advantage in the computation of the hidden layer themselves, since we are now performing a matrix multiplication of the following form:
$$ ((H\otimes M_H)W)\otimes M_O $$
where $M_H$ and $M_O$ are binary mask matrices. $M_O$ is obtained for each layer from the sampling of the policy as described in eq. \ref{Bernoulli-distrib}: each sampled action (0 or 1) is repeated so as to span the corresponding block. $M_H$ is simply the mask of the previous layer. $M_H$ and $M_O$ resemble this (here there are 3 \emph{blocks} of size 2):
\[
\begin{tikzpicture}[overlay]
  \foreach \x in {0,...,2}
    \foreach \y in {0,1,3}
      \draw (0.35+\x*1.1,0.5-\y*0.4) rectangle (1.45+\x*1.1,0.9-\y*0.4);
\end{tikzpicture}
\left(\begin{array}{cccccc}
0 & 0 & 1 & 1 & 0 & 0\\
1 & 1 & 0 & 0 & 0 & 0\\
 &  & ...\\
0 & 0 & 1 & 1 & 1 & 1
\end{array}\right)
\]
This allows us to quickly perform matrix multiplication by only considering the non-zero output elements as well as the non-zero elements in $H\otimes M_H$.

\section{Experiments}

\subsection{Model implementation}
The proposed model was implemented within Theano \citep{Bergstra2010}, a standard library for deep learning and neural networks. In addition to using optimizations offered by Theano, we also implemented specialized matrix multiplication code for the operation exposed in section \ref{sub:blockdrop}.\\
A straightforward and fairly naive CPU implementation of this operation yielded speedups of up to 5-10x, while an equally naive GPU implementation yielded speedups of up to 2-4x, both for sparsity rates of under 20\% and acceptable matrix and block sizes.\footnote{Implementations used in this paper are available at \url{http://github.com/bengioe/condnet/}}

We otherwise use fairly standard methods for our neural network. The weight matrices are initialized using the heuristic of \citet{Glorot2010}. We use a constant learning rate throughout minibatch SGD. We also use early stopping \citep{Bishop2006} to avoid overfitting. We only use fully-connected layers with $\tanh$ activations (reLu activations offer similar performance).

\subsection{Model evaluation}
We first evaluate the performance of our model on the \textbf{MNIST} digit dataset. We use a single hidden layer of 16 blocks of 16 units (256 units total), with a target sparsity rate of $\tau=6.25\%=1/16$, learning rates of $10^{-3}$ for the neural network and $5\times10^{-5}$ for the policy,  $\lambda_v=\lambda_s=200$ and $\lambda_{L2}=0.005$. Under these conditions, a test error of around 2.3\% was achieved. A normal neural network with the same number of hidden units achieves a test error of around 1.9\%, while a normal neural network with a similar \emph{amount} of computation (multiply-adds) being made (32 hidden units) achieves a test error of around 2.8\%.\\

\begin{figure}
        %\centering
  \hspace{4em}
        \begin{subfigure}[b]{0.32\textwidth}
                \includegraphics[width=\textwidth]{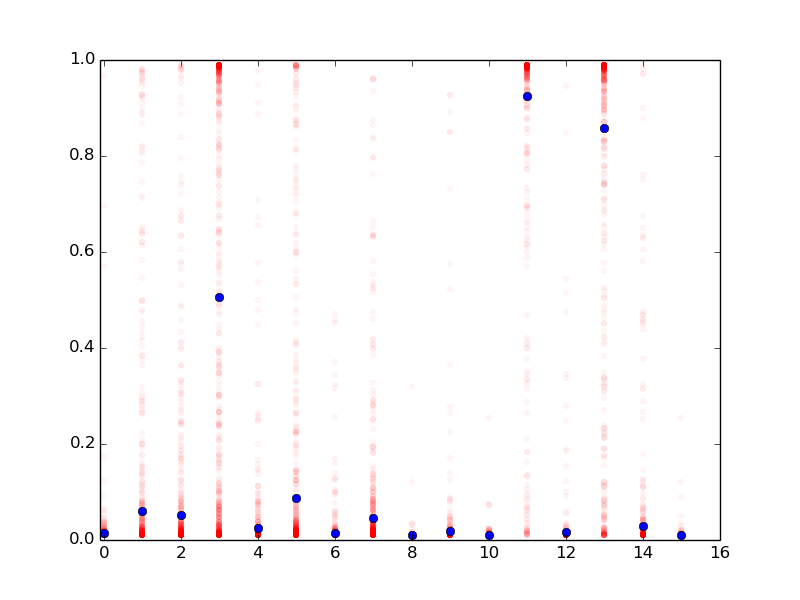}
                \caption{}
                \label{fig:mnist_probs_a}
        \end{subfigure}
        \begin{subfigure}[b]{0.32\textwidth}
                \includegraphics[width=\textwidth]{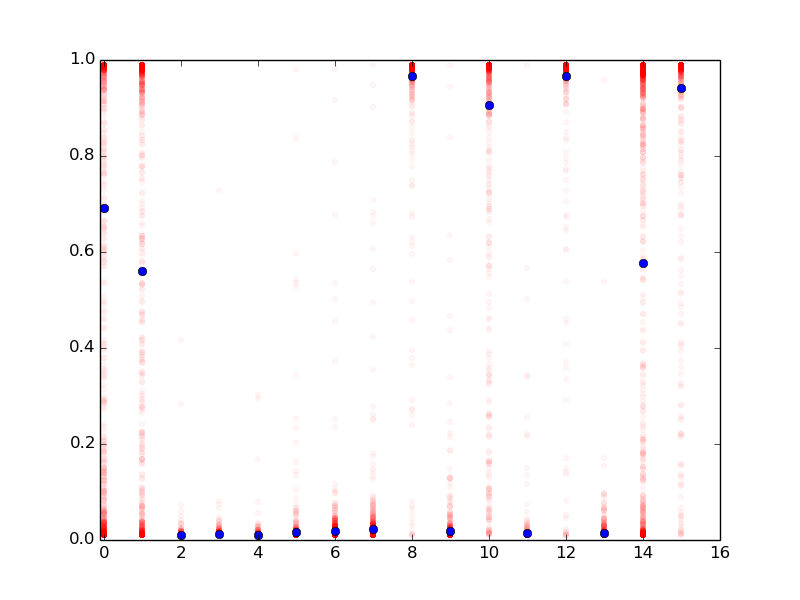}
                \caption{}
                \label{fig:mnist_probs_b}
        \end{subfigure}

        \hspace{4em}
        \begin{subfigure}[b]{0.32\textwidth}
                \includegraphics[width=\textwidth]{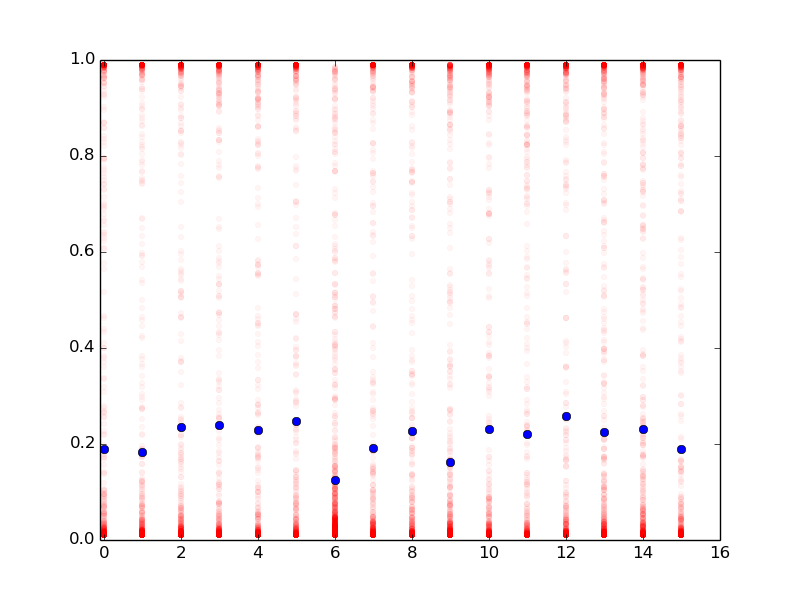}
                \caption{}
                \label{fig:mnist_probs}
        \end{subfigure}
        %~ %add desired spacing between images, e. g. ~, \quad, \qquad, \hfill etc.
        \;\;\quad
        \hspace{1em}
        \begin{subfigure}[b]{0.192\textwidth}
                \includegraphics[width=\textwidth]{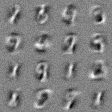}
                \;
                \caption{}
                \label{fig:sanity-2}
        \end{subfigure}
        \caption{ MNIST, (a,b,c), probability distribution of the policy, each example's probability (y axis) of activating each unit (x axis) is plotted as a transparent red dot. Redder regions represent more examples falling in the probability region. Plot (a) is for class '0', (b) for class '1', (c) for all classes. (d), weight matrix of the policy.}\label{fig:mnist}
\end{figure}

Looking at the activation of the policy (\ref{fig:mnist_probs}), we see that it tends towards what was hypothesized in section \ref{sec:sparsity}, i.e. where examples activate most units with low probability and some units with high probability. We can also observe that the policy is input-dependent in figures \ref{fig:mnist_probs_a} and \ref{fig:mnist_probs_b}, since we see different activation patterns for inputs of class '0' and inputs of class '1'.

Since the computation performed in our model is sparse, one could hope that it achieves this performance with less computation time, yet we consistently observe that models that deal with MNIST are too small to allow our specialized (\ref{sub:blockdrop}) sparse implementation to make a substantial difference. We include this result to highlight  conditions under which it is less desirable to use our model.

%./exp_6214 min valid 0.183312495403 16 [0.08722520382314913, 0.14337415068584608] [13, 8] 1.546814 2.224724 speedup 1.43826213106

Next, we consider the performance of our model on the \textbf{CIFAR-10} \citep{krizhevsky2009learning} image dataset. A brief hyperparameter search was made, and a few of the best models are shown in figure \ref{fig:cifresults}.
  These results show that it is possible to achieve similar performance with our model (denoted condnet) as with a normal neural network (denoted NN), yet using sensibly reduced computation time. A few things are worth noting; we can set $\tau$ to be lower than 1 over the number of blocks, since the model learns a policy that is actually not as sparse as $\tau$, mostly because REINFORCE pulls the policy towards higher probabilities on average. For example our best performing model has a target of $1/16$ but learns policies that average an 18\% sparsity rate (we used $\lambda_v=\lambda_s=20$, except for the first layer $\lambda_v=40$, we used $\lambda_{L2}=0.01$, and the learning rates were 0.001 for the neural net, $10^{-5}$ and $5\times 10^{-4}$ for the first and second policy layers respectively). The neural networks without conditional activations are trained with L2 regularization as well as regular unit-wise dropout.\\
We also train networks with the same architecture as our models, using blocks, but with a uniform policy (as in original dropout) instead of a learned conditional one. This model (denoted bdNN) does not perform as well as our model, showing that the dropout noise by itself is not sufficient, and that learning a policy is required to fully take benefit of this architecture.
% Preview source code for paragraph 2
\begin{figure}

  \centering
  \begin{tabular}{|c|c|c|c|c|c|c|}
\hline 
model & test error & $\tau$ & \#blocks & block size & test time & speedup \tabularnewline
\hline 
\hline 
condnet & 0.511 & 1/24 & 24,24 & 64 & 6.8s(26.2s) & 3.8$\times$\tabularnewline
\hline 
condnet & 0.514 & 1/16 & 16,32 & 16 & 1.4s (8.2s) & 5.7$\times$\tabularnewline
\hline 
condnet & \textbf{0.497} & 1/16 & 10,10 & 64 & 2.0s(10.4s) & 5.3$\times$\tabularnewline
\hline 
bdNN & 0.629 & 0.17 & 10,10 & 64 & 1.93s(10.3s) & 5.3$\times$\tabularnewline
\hline 
bdNN & 0.590 & 0.2 & 10,10 & 64 & 2.8s(10.3s) & 3.5$\times$\tabularnewline
\hline 
NN & 0.560  & - & 64,64 & 1 & 1.23s & -\tabularnewline
\hline 
NN & 0.546  & - & 128,128 & 1 & 2.31s & - \tabularnewline
\hline 
NN & \textbf{0.497} & - & 480,480 & 1 & 8.34s & -\tabularnewline
\hline 
\end{tabular}
  \caption{CIFAR-10, condnet: our approach, NN: Neural Network without the conditional activations, bdNN, block dropout Neural Network using a uniform policy. 'speedup' is how many times faster the forward pass is when using a specialized implementation (\ref{sub:blockdrop}). 'test time' is the time required to do a full pass over the test dataset using the implementation, on a \textbf{CPU}, running on a \textbf{single core}; in parenthesis is the time without the optimization. }

  \label{fig:cifresults}
\end{figure}
\begin{figure}[!h]
        \centering
%
       % ~ %add desired spacing between images, e. g. ~, \quad, \qquad, \hfill etc.
        %\begin{subfigure}[b]{0.9\textwidth}
                \includegraphics[width=\textwidth]{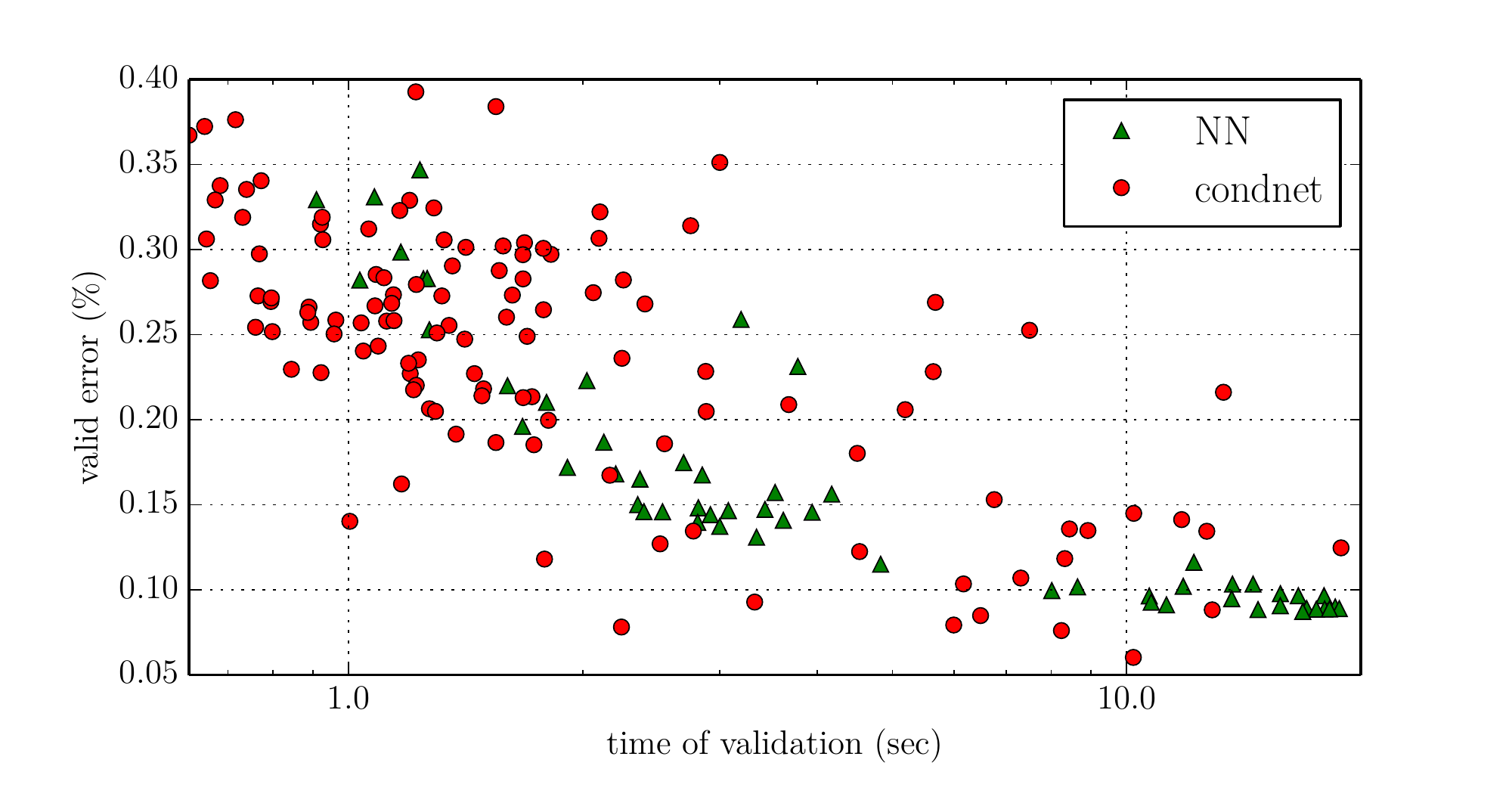}
        %\end{subfigure}
        \caption{ SVHN, each point is an experiment. The x axis is the time required
to do a full pass over the valid dataset (log scale, lower is better). Note that we plot the full hyperparameter exploration results, which is why condnet results are so varied.}
                \label{fig:svhn_valid_time}
\end{figure}

Finally we tested our model on the Street View House Numbers (\textbf{SVHN}) \citep{netzer2011reading} dataset, which also yielded encouraging results (figure \ref{fig:svhn_valid_time}). 
%Three interesting observations can be made
As we restrain the capacity of the models (by increasing sparsity or decreasing number of units), condnets retain acceptable performance with low run times, while plain neural networks suffer highly (their performance dramatically decreases with lower run times). \\
The best condnet model has a test error of 7.3\%, and runs a validation epoch in 10s (14s without speed optimization), while the best standard neural network model has a test error of 9.1\%, and runs in 16s. Note that the variance in the SVHN results (figure \ref{fig:svhn_valid_time}) is due to the mostly random hyperparameter exploration, where block size, number of blocks, $\tau$, $\lambda_v$, $\lambda_s$, as well of learning rates are randomly picked. The normal neural network results were obtained by varying the number of hidden units of a 2-hidden-layer model.

For all three datasets and all condnet models used, the required training time was higher, but still reasonable. On average experiments took 1.5 to 3 times longer (wall time).
\begin{figure}

  \centering
  \begin{tabular}{|c|c|c|c|c|c|c|}
\hline 
model & test error & $\tau$ & \#blocks & block size & test time & speedup \tabularnewline
\hline 
\hline 
condnet & 0.183 & 1/11 & 13,8 & 16 & 1.5s(2.2s) & 1.4$\times$\tabularnewline
\hline 
condnet & 0.139 & 1/25,1/7 & 27,7 & 16 & 2.8s (4.3s) & 1.6$\times$\tabularnewline
\hline 
condnet & \textbf{0.073} & 1/22 & 25,22 & 32 & 10.2s(14.1s) & 1.4$\times$\tabularnewline
\hline 
NN & 0.116 & - & 288,928 & 1 & 4.8s & -\tabularnewline
\hline 
NN & 0.100  & - & 800,736 & 1 & 10.7s & - \tabularnewline
\hline 
NN & 0.091  & - & 1280,1056 & 1 & 16.8s & -\tabularnewline
\hline 
\end{tabular}

  \caption{SVHN results (see fig \ref{fig:cifresults})}

  \label{fig:svhnresults}
\end{figure}
%Second, it seems that our current approach might limit capacity too well, as when the latter is increased condnets do worse. Third, 

\subsection{Effects of regularization}

\begin{figure}%[!b]
        \centering
%
       % ~ %add desired spacing between images, e. g. ~, \quad, \qquad, \hfill etc.
        \begin{subfigure}[b]{0.45\textwidth}
                \includegraphics[width=\textwidth]{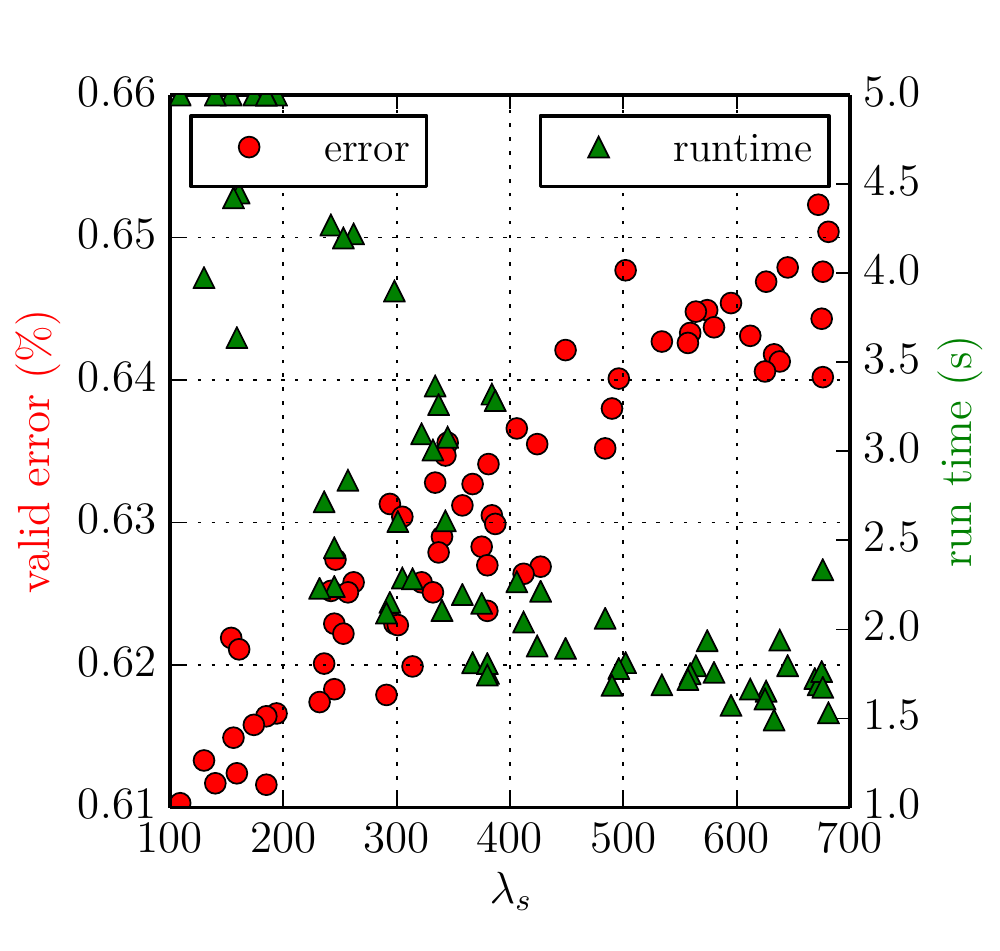}
                \caption{}
                \label{fig:cifar_lambda_s}
        \end{subfigure}
  %\hspace{-1.8em}
        \begin{subfigure}[b]{0.45\textwidth}
                \includegraphics[width=\textwidth]{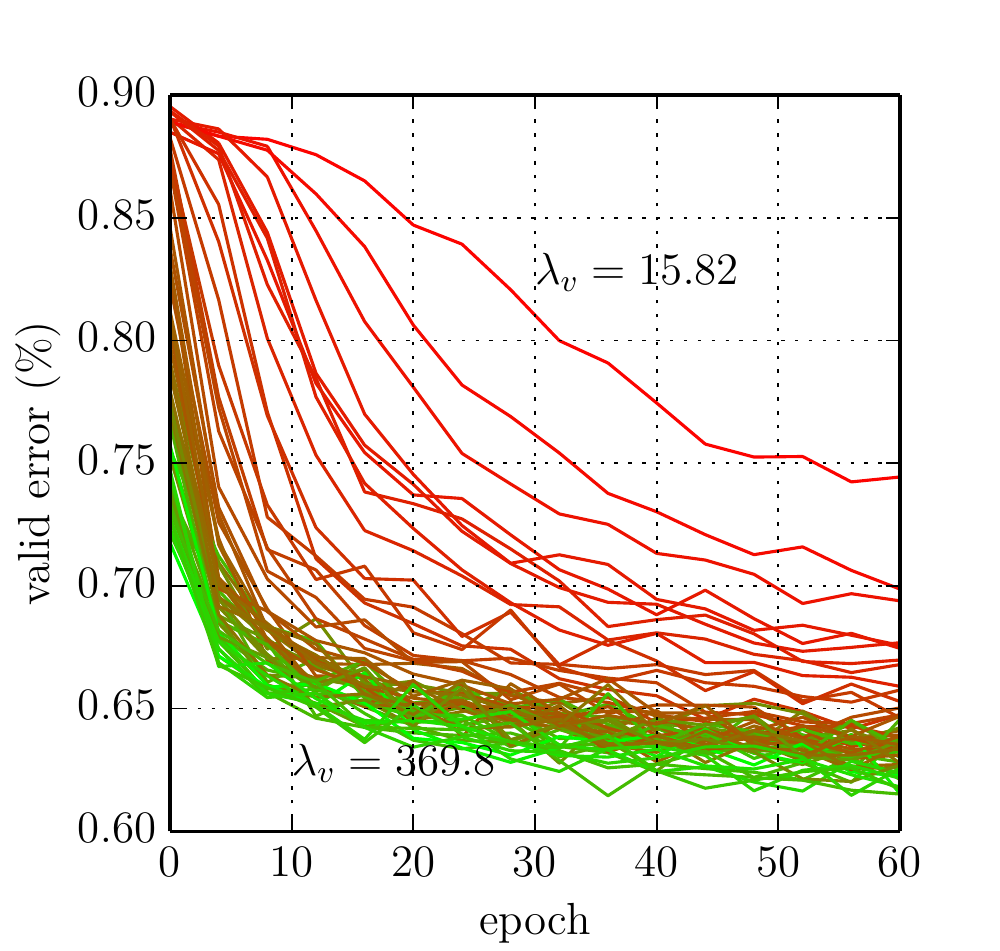}
                \caption{}
                \label{fig:cifar_lambda_v}
        \end{subfigure}
        \caption{ CIFAR-10, (a) each pair of circle and triangle is an experiment made with a given lambda (x axis), resulting in a model with a certain error and running time (y axes). As $\lambda_s$ increases the running time decreases, but so does performance. (b) The same model is being trained with different values of $\lambda_v$. Redder means lower, greener means higher.}
\end{figure}

The added regularization proposed in section \ref{sec:sparsity} seems to play an important role in our ability to train the conditional model. When using only the prediction score, we observed that the algorithm tried to compensate by recruiting more units and saturating their participation probability, or even failed by dismissing very early what were probably considered \emph{bad} units. \\
In practice, the variance regularization term $L_v$ only slightly affects the prediction accuracy and learned policies of models, but we have observed that it significantly speeds up the training process, probably by encouraging policies to become less uniform earlier in the learning process. This can be seen in figure \ref{fig:cifar_lambda_v}, where we train a model with different values of $\lambda_v$. When $\lambda_v$ is increased, the first few epochs have a much lower error rate.

It is possible to tune some hyperparameters to affect the point at which the trade-off between computation speed and performance lies, thus one could push the error downwards at the expense of also more computation time. This is suggested by figure \ref{fig:cifar_lambda_s}, which shows the effect of one such hyperparameter ($\lambda_s$) on both running times and performance for the CIFAR dataset.  Here it seems that $\lambda \sim [300,400]$ offers the best trade-off, yet other values could be selected, depending on the specific requirements of an application.

\section{Related work}
\citet{Ba2013} proposed a learning algorithm called \textit{standout} for computing an
input-dependent dropout distribution at every node. As opposed to our layer-wise method, standout
computes a one-shot dropout mask over the entire network, conditioned on the input to the
network. Additionally, masks are unit-wise, while our approach uses masks that span blocks of units.
\citet{bengio2013estimating} introduced Stochastic Times Smooth neurons as gaters for conditional
computation within a deep neural network. STS neurons are highly non-linear and non-differentiable
functions learned using estimators of the gradient obtained through REINFORCE. They allow a sparse
binary gater to be computed as a function of the input, thus reducing computations in the
then sparse activation of hidden layers.

\citet{Stollenga2014} recently proposed to learn a sequential decision process over the
filters of a convolutional neural network (CNN). As in our work, a direct policy search
method was chosen to find the parameters of a control policy. Their problem formulation
differs from ours mainly in the notion of decision ``stage''. In their model, an input is
first fed through a network, the activations are computed during forward propagation then
they are served to the next decision stage. The goal of the policy is to select relevant
filters from the previous stage so as to improve the decision accuracy on the current
example.  They also use a gradient-free evolutionary algorithm, in contrast to our gradient-based method.
%\citet{Schaul2011}

The Deep Sequential Neural Network (DSNN) model of \citet{Denoyer2014} is possibly closest
to our approach. The control process is carried over the layers of the network and uses
the output of the previous layer to compute actions. The REINFORCE algorithm is used to
train the policy with the reward/cost function being defined as the loss at the output in
the base network. DSNN considers the general problem of choosing between between different
type of mappings (weights) in a composition of functions. However, they test their model
on datasets in which different modes are proeminent, making it easy for a policy to
distinguish between them.

Another point of comparison for our work are attention models \citep{Mnih2014b, gregor2015draw, xu2015show}. These models typically learn a policy, or a form of policy, that allows them to selectively attend to parts of their input \emph{sequentially}, in a visual 2D environnement. Both attention and our approach aim to reduce computation times. While attention aims to perform dense computations on subsets of the inputs, our approach aims to be more general, since the policy focuses on subsets of the whole computation (it is in a sense more distributed). It should also be possible to combine these approaches, since one acts on the input space and the other acts on the representation space, altough the resulting policies would be much more complex, and not necessarily easily trainable.
%Yet, it is not unthinkable that both approaches could be combined.

\section{Conclusion}

This paper presents a method for tackling the problem of conditional
computation in deep networks by using reinforcement learning. We
propose a type of parameterized conditional computation policy that maps the
activations of a layer to a Bernoulli mask.  The reinforcement signal
accounts for the loss function of the network in its prediction task,
while the policy network itself is regularized to account for the
desire to have sparse computations.  The REINFORCE algorithm is used
to train policies to optimize this cost.  Our experiments show that it
is possible to train such models at the same levels of accuracy as
their standard counterparts. Additionally, it seems possible to
execute these similarly accurate models faster due to their sparsity.
Furthermore, the model has a few simple parameters that allow to
control the trade-off between accuracy and running time.

The use of REINFORCE could be replaced by a more efficient policy
search algorithm, and also, perhaps, one in which rewards (or costs)
as described above are replaced by a more sequential variant.  The
more direct use of computation time as a cost may prove
beneficial.  In general, we consider conditional computation to be an
area in which reinforcement learning could be very useful, and
deserves further study.

All the running times reported in the Experiments section are for a CPU, running on a single core. The motivation for this is to explore deployment of large neural networks on cheap, low-power, single core CPUs such as phones, while retaining high model capacity and expressiveness.  While the results presented here show that our model for conditional computation can achieve speedups in this context, it is worth also investigating adaptation of these sparse computation models in multicore/GPU architectures; this is the subject of ongoing work.

\section*{Acknowledgements}

The authors gratefully acknowledge financial support for this work by the Samsung Advanced Institute of Technology (SAIT), the Natural Sciences and Engineering Research Council of Canada (NSERC) and the Fonds de recherche du Qu\'{e}bec - Nature et Technologies (FQRNT).

%\printbibliography

\bibliography{library}

\begin{thebibliography}{24}
\providecommand{\natexlab}[1]{#1}
\providecommand{\url}[1]{\texttt{#1}}
\expandafter\ifx\csname urlstyle\endcsname\relax
  \providecommand{\doi}[1]{doi: #1}\else
  \providecommand{\doi}{doi: \begingroup \urlstyle{rm}\Url}\fi

\bibitem[Ba \& Frey(2013)Ba and Frey]{Ba2013}
Ba, Jimmy and Frey, Brendan.
\newblock Adaptive dropout for training deep neural networks.
\newblock In Burges, C.J.C., Bottou, L., Welling, M., Ghahramani, Z., and
  Weinberger, K.Q. (eds.), \emph{Advances in Neural Information Processing
  Systems 26}, pp.\  3084--3092. Curran Associates, Inc., 2013.
\newblock URL
  \url{http://papers.nips.cc/paper/5032-adaptive-dropout-for-training-deep-neural-networks.pdf}.

\bibitem[Bengio et~al.(1994)Bengio, Simard, and Frasconi]{Bengio-et-al-TNN1994}
Bengio, Y., Simard, P., and Frasconi, P.
\newblock Learning long-term dependencies with gradient descent is difficult.
\newblock \emph{IEEE Transactions on Neural Nets}, pp.\  157--166, 1994.

\bibitem[Bengio et~al.(2013)Bengio, L{\'e}onard, and
  Courville]{bengio2013estimating}
Bengio, Yoshua, L{\'e}onard, Nicholas, and Courville, Aaron.
\newblock Estimating or propagating gradients through stochastic neurons for
  conditional computation.
\newblock \emph{arXiv preprint arXiv:1308.3432}, 2013.

\bibitem[Bergstra et~al.(2010)Bergstra, Breuleux, Bastien, Lamblin, Pascanu,
  Desjardins, Turian, Warde-Farley, and Bengio]{Bergstra2010}
Bergstra, James, Breuleux, Olivier, Bastien, Fr{\'{e}}d{\'{e}}ric, Lamblin,
  Pascal, Pascanu, Razvan, Desjardins, Guillaume, Turian, Joseph, Warde-Farley,
  David, and Bengio, Yoshua.
\newblock Theano: a {CPU} and {GPU} math expression compiler.
\newblock In \emph{Proceedings of the Python for Scientific Computing
  Conference ({SciPy})}, June 2010.
\newblock Oral Presentation.

\bibitem[Bishop(2006)]{Bishop2006}
Bishop, Christopher~M.
\newblock \emph{Pattern Recognition and Machine Learning (Information Science
  and Statistics)}.
\newblock Springer-Verlag New York, Inc., Secaucus, NJ, USA, 2006.
\newblock ISBN 0387310738.

\bibitem[Davis \& Arel(2013)Davis and Arel]{davis2013low}
Davis, Andrew and Arel, Itamar.
\newblock Low-rank approximations for conditional feedforward computation in
  deep neural networks.
\newblock \emph{arXiv preprint arXiv:1312.4461}, 2013.

\bibitem[Deisenroth et~al.(2013)Deisenroth, Neumann, and
  Peters]{Deisenroth2013}
Deisenroth, Marc~Peter, Neumann, Gerhard, and Peters, Jan.
\newblock A survey on policy search for robotics.
\newblock \emph{Foundations and Trends in Robotics}, 2\penalty0 (1-2):\penalty0
  1--142, 2013.
\newblock \doi{10.1561/2300000021}.
\newblock URL \url{http://dx.doi.org/10.1561/2300000021}.

\bibitem[Denoyer \& Gallinari(2014)Denoyer and Gallinari]{Denoyer2014}
Denoyer, Ludovic and Gallinari, Patrick.
\newblock Deep sequential neural network.
\newblock \emph{CoRR}, abs/1410.0510, 2014.
\newblock URL \url{http://arxiv.org/abs/1410.0510}.

\bibitem[Glorot \& Bengio(2010)Glorot and Bengio]{Glorot2010}
Glorot, Xavier and Bengio, Yoshua.
\newblock Understanding the difficulty of training deep feedforward neural
  networks.
\newblock In \emph{Proceedings of the Thirteenth International Conference on
  Artificial Intelligence and Statistics, {AISTATS} 2010, Chia Laguna Resort,
  Sardinia, Italy, May 13-15, 2010}, pp.\  249--256, 2010.
\newblock URL \url{http://www.jmlr.org/proceedings/papers/v9/glorot10a.html}.

\bibitem[Gregor et~al.(2015)Gregor, Danihelka, Graves, and
  Wierstra]{gregor2015draw}
Gregor, Karol, Danihelka, Ivo, Graves, Alex, and Wierstra, Daan.
\newblock Draw: A recurrent neural network for image generation.
\newblock \emph{arXiv preprint arXiv:1502.04623}, 2015.

\bibitem[He et~al.(2015)He, Zhang, Ren, and Sun]{he2015delving}
He, Kaiming, Zhang, Xiangyu, Ren, Shaoqing, and Sun, Jian.
\newblock Delving deep into rectifiers: Surpassing human-level performance on
  imagenet classification.
\newblock \emph{arXiv preprint arXiv:1502.01852}, 2015.

\bibitem[Hinton et~al.(2012)Hinton, Srivastava, Krizhevsky, Sutskever, and
  Salakhutdinov]{Hinton2012}
Hinton, Geoffrey~E., Srivastava, Nitish, Krizhevsky, Alex, Sutskever, Ilya, and
  Salakhutdinov, Ruslan.
\newblock Improving neural networks by preventing co-adaptation of feature
  detectors.
\newblock \emph{CoRR}, abs/1207.0580, 2012.
\newblock URL \url{http://arxiv.org/abs/1207.0580}.

\bibitem[Hochreiter(1991)]{Hochreiter91-small}
Hochreiter, S.
\newblock { Untersuchungen zu dynamischen neuronalen Netzen. Diploma thesis,
  T.U. M\"{u}nich}, 1991.

\bibitem[Krizhevsky \& Hinton(2009)Krizhevsky and
  Hinton]{krizhevsky2009learning}
Krizhevsky, Alex and Hinton, Geoffrey.
\newblock Learning multiple layers of features from tiny images, 2009.

\bibitem[Martens(2010)]{Martens2010}
Martens, James.
\newblock Deep learning via hessian-free optimization.
\newblock In \emph{Proceedings of the 27th International Conference on Machine
  Learning (ICML-10), June 21-24, 2010, Haifa, Israel}, pp.\  735--742, 2010.
\newblock URL \url{http://www.icml2010.org/papers/458.pdf}.

\bibitem[Mnih et~al.(2014)Mnih, Heess, Graves, and kavukcuoglu]{Mnih2014b}
Mnih, Volodymyr, Heess, Nicolas, Graves, Alex, and kavukcuoglu, koray.
\newblock Recurrent models of visual attention.
\newblock In Ghahramani, Z., Welling, M., Cortes, C., Lawrence, N.D., and
  Weinberger, K.Q. (eds.), \emph{Advances in Neural Information Processing
  Systems 27}, pp.\  2204--2212. Curran Associates, Inc., 2014.
\newblock URL
  \url{http://papers.nips.cc/paper/5542-recurrent-models-of-visual-attention.pdf}.

\bibitem[Netzer et~al.(2011)Netzer, Wang, Coates, Bissacco, Wu, and
  Ng]{netzer2011reading}
Netzer, Yuval, Wang, Tao, Coates, Adam, Bissacco, Alessandro, Wu, Bo, and Ng,
  Andrew~Y.
\newblock Reading digits in natural images with unsupervised feature learning.
\newblock In \emph{NIPS workshop on deep learning and unsupervised feature
  learning}, volume 2011, pp.\ ~5. Granada, Spain, 2011.

\bibitem[Pearlmutter(1994)]{Pearlmutter1994}
Pearlmutter, Barak~A.
\newblock Fast exact multiplication by the hessian.
\newblock \emph{Neural Comput.}, 6\penalty0 (1):\penalty0 147--160, January
  1994.
\newblock ISSN 0899-7667.
\newblock \doi{10.1162/neco.1994.6.1.147}.
\newblock URL \url{http://dx.doi.org/10.1162/neco.1994.6.1.147}.

\bibitem[Puterman(1994)]{Puterman1994}
Puterman, Martin~L.
\newblock \emph{Markov Decision Processes: Discrete Stochastic Dynamic
  Programming}.
\newblock John Wiley \& Sons, Inc., New York, NY, USA, 1st edition, 1994.
\newblock ISBN 0471619779.

\bibitem[Rumelhart et~al.(1988)Rumelhart, Hinton, and
  Williams]{rumelhart1988learning}
Rumelhart, David~E, Hinton, Geoffrey~E, and Williams, Ronald~J.
\newblock Learning representations by back-propagating errors.
\newblock \emph{Cognitive modeling}, 5, 1988.

\bibitem[Silver et~al.(2014)Silver, Lever, Heess, Degris, Wierstra, and
  Riedmiller]{Silver2014}
Silver, David, Lever, Guy, Heess, Nicolas, Degris, Thomas, Wierstra, Daan, and
  Riedmiller, Martin.
\newblock Deterministic policy gradient algorithms.
\newblock In \emph{Proceedings of the 31th International Conference on Machine
  Learning, {ICML} 2014, Beijing, China, 21-26 June 2014}, pp.\  387--395,
  2014.
\newblock URL \url{http://jmlr.org/proceedings/papers/v32/silver14.html}.

\bibitem[Stollenga et~al.(2014)Stollenga, Masci, Gomez, and
  Schmidhuber]{Stollenga2014}
Stollenga, Marijn~F, Masci, Jonathan, Gomez, Faustino, and Schmidhuber,
  J\"{u}rgen.
\newblock Deep networks with internal selective attention through feedback
  connections.
\newblock In Ghahramani, Z., Welling, M., Cortes, C., Lawrence, N.D., and
  Weinberger, K.Q. (eds.), \emph{Advances in Neural Information Processing
  Systems 27}, pp.\  3545--3553. Curran Associates, Inc., 2014.
\newblock URL
  \url{http://papers.nips.cc/paper/5276-deep-networks-with-internal-selective-attention-through-feedback-connections.pdf}.

\bibitem[Williams(1992)]{Williams1992}
Williams, Ronald~J.
\newblock Simple statistical gradient-following algorithms for connectionist
  reinforcement learning.
\newblock \emph{Machine Learning}, 8\penalty0 (3-4):\penalty0 229--256, 1992.
\newblock ISSN 0885-6125.
\newblock \doi{10.1007/BF00992696}.
\newblock URL \url{http://dx.doi.org/10.1007/BF00992696}.

\bibitem[Xu et~al.(2015)Xu, Ba, Kiros, Courville, Salakhutdinov, Zemel, and
  Bengio]{xu2015show}
Xu, Kelvin, Ba, Jimmy, Kiros, Ryan, Courville, Aaron, Salakhutdinov, Ruslan,
  Zemel, Richard, and Bengio, Yoshua.
\newblock Show, attend and tell: Neural image caption generation with visual
  attention.
\newblock \emph{arXiv preprint arXiv:1502.03044}, 2015.

\end{thebibliography}
\bibliographystyle{iclr2016_conference}

\newpage
\appendix
\section{Algorithm}
\label{algorithm-pseudocode}
The forward pass in our model is done as described in the algorithm below (\ref{alg:forward1}), both at train time and test time.

\label{algorithm-pseudocode}
\begin{algorithm}[H]
  \SetKwInOut{Input}{input}
  \Input{$\mathbf{x}$}
  $\mathbf{h}_0 \leftarrow \mathbf{x}$\\
  $\mathbf{u}_0 \leftarrow \mathbf{1}$ \tcp*{the input mask is ones}
  \For{each hidden layer $l\in{1,...,L}$}{
    $\mathbf{p}_l \leftarrow \mbox{sigm}(\mathbf{Z}^{(l)}\mathbf{h}_{l-1}+\mathbf{d}^{(l)}) = \pi_l(\mathbf{u}_l|\mathbf{s}_l=\mathbf{h}_{l-1})$ \\
    $\mathbf{u}_l \sim \mbox{Ber}(\mathbf{p}_l)$ \tcp*{sample Bernoulli from probablities $p_l$}
    \If{$blocksize$ $> 1$}{
      extend $\mathbf{u}_l$ by repeating each value $blocksize$ times
    }
    \tcp{this operation can be performed efficiently as described in section \ref{sub:blockdrop}:}
    $\mathbf{h}_l \leftarrow f\left(\mathbf{W}^{(l)}(\mathbf{h}_{l-1}\otimes\mathbf{u}_{l-1})+\mathbf{b}^{(l)}\right) \otimes \mathbf{u}_l$ 
  }
  \caption{Single-input forward pass}
  \label{alg:forward1}
\end{algorithm}
This algorithm can easily be extended to the minibatch setting by replacing vector operations by matrix operations. Note that in the case of classification, the last layer is a softmax layer and is not multiplied by a mask.

\begin{algorithm}[H]
  \SetKwInOut{Input}{input}
  \Input{$\mathbf{x}$}
  %$\mathbf{m_o} \leftarrow \mathbf{1}$ \tcp*{the input mask is ones}
  $\hat{\mathbf{y}} = \mbox{forward}(\mathbf{x})$ \tcp*{given the output of the forward pass}
  $c \leftarrow C(\mathbf{x}) = -\log P(\hat{\mathbf{y}}|\mathbf{x})$\\
  $\mathcal{L} \leftarrow c +  \lambda_s (L_b + L_e) + \lambda_v (L_v) + \lambda_{L2}\|\Theta_{NN}\|^2 + \lambda_{L2}\|\Theta_\pi\|^2$ \tcp*{as in sections \ref{sec:sparsity} and \ref{sec:algo}}
  \tcp{update the neural network weights:}
  $\Theta_{NN} \leftarrow \Theta_{NN} - \alpha \nabla_{\Theta_{NN}} \mathcal{L}$\\
  \tcp{update the policy weights:}
  \For{each hidden layer $l\in{1,...,L}$}{
    $\theta_{l} \leftarrow \theta_{l} - \alpha_\pi \underbrace{ c \nabla_{\theta_{l}} \log \mathbf{p}_l}_{\small\mbox{REINFORCE}} - \alpha \nabla_{\theta_{l}} \mathcal{L}$ \tcp*{where $\mathbf{p}_l$ is computed as in algorithm \ref{alg:forward1}}
  }
  \caption{Single-input backward pass}
  \label{alg:backward1}
\end{algorithm}
Note that in line 4, some gradients are zeroes, for example the gradient of the L2 regularisation of $\Theta_\pi$ with respect to $\Theta_{NN}$ is zero. Similarly in line 5, the gradient of $c$ with respect to $\Theta_\pi$ is zero, which is why we have to use REINFORCE to approximate a gradient in the direction that minimizes $c$.

This algorithm can be extended to the minibatch setting efficiently by replacing the gradient computations in line 7 with the use of the so called R-op, as described in section \ref{Rop}, and other computations as is usually done in the minibatch setting with matrix operations.
\section{REINFORCE}
\label{REINFORCE}
REINFORCE \citep{Williams1992}, also known as the likelihood-ratio method, is a policy search algorithm. It aims to use gradient methods to improve a given parameterized policy.

In reinforcement learning, a sequence of state-action-reward tuples is described as a trajectory $\tau$ . The objective function of a parameterized policy $\pi_\theta$ for the cumulative return of a trajectory $\tau$ is described as:
\begin{equation*} J(\theta) = \mathbb{E}_\tau^{\pi_\theta}\left\{ \sum_{t=1}^T r(S_t,A_t|S_0=s_0)\right\}
\end{equation*}
where $s_0$ is the initial state of the trajectory. Let $R(\tau)$ denote the return for trajectory $\tau$. The gradient of the objective with respect to the parameters of the policy is:
\begin{align}
  \nabla_\theta J(\theta) &=\nabla_\theta \mathbb{E}_\tau^{\pi_\theta} \{R(\tau)\} \notag \\
  &= \nabla_\theta \int_\tau \mathbb{P}\{\tau|\theta\}R(\tau)d\tau \notag\\
  &= \int_\tau \nabla_\theta \left[\mathbb{P}\{\tau|\theta\} R(\tau)\right] d\tau \label{exchange}
\end{align}
Note that the interchange in \eqref{exchange} is only valid under some assumptions (see \cite{Silver2014}).
\begin{align}
  \nabla_\theta J(\theta) &=  \int_\tau \nabla_\theta
     \left[\mathbb{P}\{\tau|\theta\} R(\tau)\right] d\tau \notag \\
  &= \int_\tau \left[R(\tau)\nabla_\theta \mathbb{P}\{\tau|\theta\} + \nabla_\theta R(\tau)\mathbb{P}\{\tau|\theta\}\right] d\tau \label{productrule}\\
  &= \int_\tau \left[\frac{R(\tau)}{\mathbb{P}\{\tau|\theta\}} \nabla_\theta \mathbb{P}\{\tau|\theta\} + \nabla_\theta R(\tau)\right] \mathbb{P}\{\tau|\theta\} d\tau \notag \\    
  &= \mathbb{E}_\tau^{\pi_\theta}\left\{R(\tau)\nabla_\theta \log\mathbb{P}\{\tau|\theta\}+\nabla_\theta R(\tau)\right\} \label{logrule}
\end{align}
The product rule of derivatives is used in \eqref{productrule}, and the derivative of a $\log$ in \eqref{logrule}. Since $R(\tau)$ does not depend on $\theta$ directly, the gradient $\nabla_\theta R(\tau)$ is zero. We end up with this gradient:
\begin{align}
\nabla_\theta J(\theta) &= \mathbb{E}_\tau^{\pi_\theta}\left\{R(\tau)\nabla_\theta \log\mathbb{P}\{\tau|\theta\}\right\}
\end{align}
Without knowing the transition probabilities, we cannot compute the probability of our trajectories $\mathbb{P}\{\tau|\theta\}$, or their gradient. Fortunately we are in a MDP setting, and we can make use of the Markov property of the trajectories to compute the gradient:
\begin{align}
  \nabla_\theta \log\mathbb{P}\{\tau|\theta\} &=\nabla_\theta \log \left[ p(s_0)\prod_{t=1}^T\mathbb{P}\{s_{t+1}|s_t,a_t\}\pi_\theta(a_t|s_t)\right] \notag\\
  &= \nabla_\theta \log p(s_0) + \sum_{t=1}^T \nabla_\theta \log \mathbb{P}\{s_{t+1}|s_t,a_t\} + \nabla_\theta \log \pi_\theta(a_t|s_t) \label{eq:gradiszero}\\
  &= \sum_{t=1}^T\nabla_\theta \log\pi_\theta(a_t|s_t) \notag
\end{align}
In \eqref{eq:gradiszero}, $p(s_0)$ does not depend on $\theta$, so the gradient is zero. Similarly, $\mathbb{P}\{s_{t+1}|s_t,a_t\}$ does not depend on $\theta$ (not directly at least), so the gradient is also zero.
We end up with the gradient of the log policy, which is easy to compute.

In our particular case, the trajectories only have a single step and the reward of the trajectory is the neural network cost $C(\mathbf{x})$, thus the summation dissapears and the gradient found in \eqref{eq:grad-loss-sigm} is found by taking the log of the probability of our Bernoulli sample:
\begin{align}
  \nabla_{\theta_l} C(\mathbf{x}) &= \mathbb{E}\left\{C(\mathbf{x}) \nabla_{\theta_l} \log \pi_{\theta_l} (\mathbf{u}|\mathbf{s}) \right\} \notag \\
  &= \mathbb{E}\left\{C(\mathbf{x}) \nabla_{\theta_l} \log \prod_{i=1}^k \sigma_i^{u_i} (1 - \sigma_i)^{(1 - u_i)} \right\} \notag \\
  &= \mathbb{E}\left\{C(\mathbf{x}) \nabla_{\theta_l} \sum_{i=1}^k\log \left[\sigma_i u_i + (1 - \sigma_i)(1 - u_i)\right] \right\} \notag
\end{align}

\end{document}